\documentclass[letterpaper, 10 pt, conference]{ieeeconf}  
\IEEEoverridecommandlockouts                              
\usepackage{graphics}    
\usepackage{times}       
\usepackage{amsmath}     
\usepackage{amssymb}     
\usepackage{graphicx}
\usepackage{algorithm}
\usepackage[colorlinks=true,linkcolor=black,citecolor=black,urlcolor=blue]{hyperref}
\usepackage{flushend}
\usepackage{glossaries}
\usepackage[noend]{algpseudocode}
\usepackage{booktabs} 
\usepackage{tabularx} 
\usepackage{adjustbox}

\usepackage[font=footnotesize]{caption}
\usepackage[font=footnotesize]{subcaption}

\def\secref#1{Sec.~\ref{#1}}
\def\figref#1{Fig.~\ref{#1}}
\def\tabref#1{Tab.~\ref{#1}}
\def\eqref#1{Eq.~(\ref{#1})}

\def\ie{{i.e.}}

\def\lidar{LiDAR}
\def\lidars{LiDARs}

\def\rgb{RGB}
\def\gnss{GNSS}

\usepackage{amsopn}

\newcommand{\bv}{\mathbf{v}}

\newcommand{\bt}{\mathbf{t}}

\newcommand{\bX}{\mathbf{X}}
\newcommand{\bZ}{\mathbf{Z}}
\newcommand{\bR}{\mathbf{R}}

\newcommand{\bbSE}{\mathbb{SE}}

\newcommand{\bn}{\mathbf{n}}

\newcommand{\bp}{\mathbf{p}}

\newcommand{\tTov}{\mathrm{log}}

\newcommand{\bOmega}{\mathbf{\Omega}}

\DeclareMathOperator*{\argmin}{argmin}

\def\g2o{$g^2o$}
\def\t2v{\mathrm{log}}
\def\v2t{\mathrm{exp}}
\def\ev2t{\mathrm{ev2t}}

\newacronym{slam}{SLAM}{Simultaneous Localization and Mapping}
\newacronym{sfm}{SfM}{Structure from Motion}
\newacronym{pgo}{PGO}{Pose-Graph Optimization}
\newacronym{vpr}{VPR}{Visual Place Recognition}
\newacronym{sgd}{SGD}{Stochastic Gradient Descent}
\newacronym{ils}{ILS}{Iterative Least-Squares}
\newacronym{icp}{ICP}{Iterative Corresponding Point}
\newacronym{gn}{GN}{Gauss-Newton}
\newacronym{lm}{LM}{Levenberg-Marquardt}
\newacronym{pcg}{PCG}{Preconditioned Conjugate Gradient}
\newacronym{map}{MAP}{Maximum-A-Posteriori}
\newacronym{gf}{GF}{Gaussian Filters}
\newacronym{pf}{PF}{Particle Filters}
\newacronym{sdp}{SDP}{Semi-Definite Programming}
\newacronym{bst}{BST}{Binary Search Tree}
\newacronym{ndt}{NDT}{Normal Distributed Transform}
\newacronym{ba}{BA}{Bundle Adjustment}
\newacronym{imu}{IMU}{Inertial Measurement Unit}
\newacronym{pps}{PPS}{Pulse Per Second}	

\def\secref#1{Sec.~\ref{#1}}
\def\figref#1{Fig.~\ref{#1}}
\def\tabref#1{Tab.~\ref{#1}}
\def\eqref#1{Eq.~(\ref{#1})}

\usepackage{lipsum}

\title{\LARGE \bf VBR: A Vision Benchmark in Rome} 

\author{
	\large
	Leonardo Brizi* \quad\quad
	Emanuele Giacomini* \quad\quad
	Luca Di Giammarino \quad\quad
	Simone Ferrari\\\\
	\centering
	Omar Salem \quad\quad
        Lorenzo De Rebotti \quad\quad
	Giorgio Grisetti
\thanks{All authors are with the Department of Computer, Control, and Management Engineering ``Antonio Ruberti", Sapienza University of Rome, Italy,
Email:\,\,{\tt\footnotesize{\{brizi, giacomini, digiammarino, s.ferrari, salem, derebotti, grisetti\}@diag.uniroma1.it.}}}%
\thanks{This work has been accepted for publication in the IEEE International Conference on Robotics and Automation (ICRA) 2024.}
\thanks{This work has been supported by PNRR MUR project PE0000013-FAIR.}
\thanks{*The authors contributed equally.}
}

\begin{document}
\maketitle
\thispagestyle{empty}
\pagestyle{empty}

\begin{abstract}
  

 
This paper presents a vision and perception research dataset collected in Rome, featuring RGB data, 3D point clouds, IMU, and GPS data. We introduce a new benchmark targeting visual odometry and SLAM, to advance the research in autonomous robotics and computer vision. This work complements existing datasets by simultaneously addressing several issues, such as environment diversity, motion patterns, and sensor frequency. It uses up-to-date devices and presents effective procedures to accurately calibrate the intrinsic and extrinsic of the sensors while addressing temporal synchronization. During recording, we cover multi-floor buildings, gardens, urban and highway scenarios. Combining handheld and car-based data collections, our setup can simulate any robot (quadrupeds, quadrotors, autonomous vehicles). The dataset includes an accurate 6-dof ground truth based on a novel methodology that refines the RTK-GPS estimate with \lidar~ point clouds through \gls{ba}. All sequences divided in training and testing are accessible at \href{https://rvp-group.net/datasets/slam.html}{www.rvp-group.net/datasets/slam}. 
\end{abstract}

\section{Introduction}
\label{sec:intro}
Computer vision communities have relied on standard datasets to enhance their techniques since the early days. When ground truth data was accessible for a specific task, these communities devised appropriate metrics to evaluate the accuracy of their algorithm's results. With the rapid advancement of machine learning, datasets equipped with ground truth have become essential inputs for algorithms designed to learn intricate, non-parametric models.
After KITTI, several other multi-sensor datasets \cite{maddern20171, schops2019bad, kim2020mulran, zhang2022hilti} have been presented, but no one seemed to have the same impact on the robotics and computer vision community as the original work \cite{geiger2012we}.

Whereas the merits of KITTI are undisputed, and the core ideas are still valid, the dataset shows its years. The available sensors in the last decade improved significantly, and the same holds for computing devices and ground truth systems. Perhaps the main shortcoming of many datasets \cite{geiger2012we,kim2020mulran,eu_longterm_dataset,boras} is the limited positional ground truth that is purely based on RTK-GPS and IMU and suffers from synchronization issues. In addition to that, the work is targeted at autonomous driving, hence, the data cover only road-like scenarios.  

Other works aimed at addressing other aspects, such as seasonal changes~\cite{maddern20171}, offering hand-held motion with a more accurate ground truth~\cite{ramezani2020newer}. Still, to our knowledge, none of the recent datasets seem to address multiple issues.
\begin{figure}
\vspace{0.5cm}
  \centering
  \includegraphics[width=0.99\linewidth]{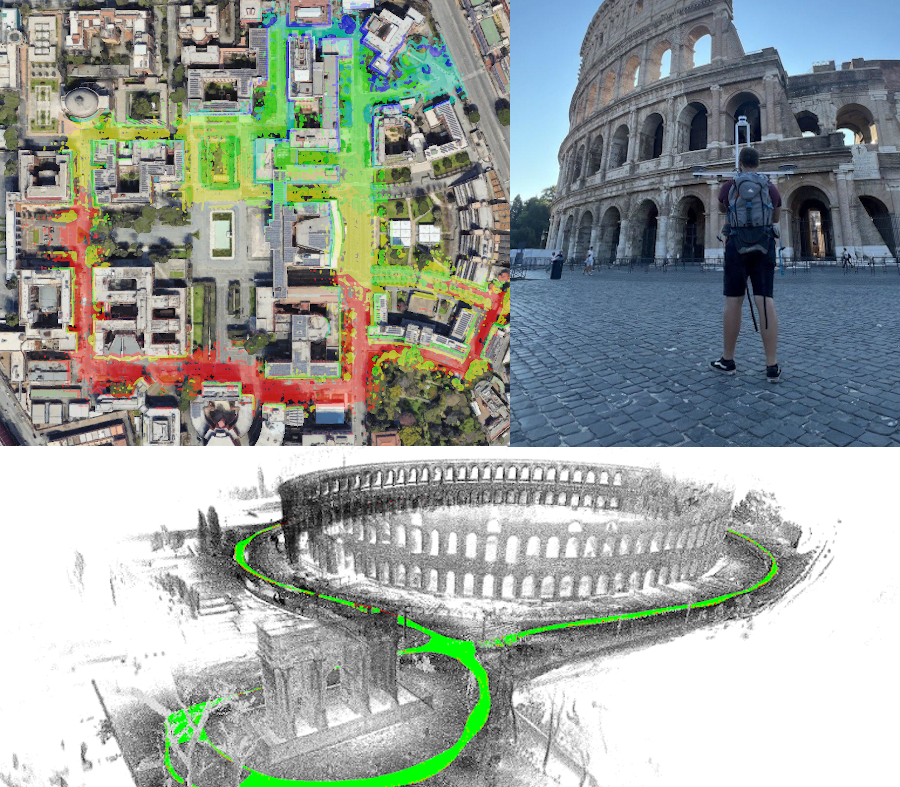}
  \caption{A summary of our dataset. Data illustrating some of the sequences recorded (top). 3D mapping done with of our ground truth (bottom).}
  \label{fig:3dCar}
\end{figure}
In this work, we present a contribution that aims to approach all these aspects simultaneously. At the moment of writing, we propose 6 datasets acquired with a hardware-synchronized sensor setting consisting of a 3D \lidar, a stereo camera with a large baseline, an RTK-GPS, and an inertial sensor. Our data covers some of the most characteristic areas of Rome, spanning over 40 km of trajectory in almost 4 hours of recording. The raw data have a footprint of about 2TB.  The sequences have been recorded in different environments, covering urban, forest, and indoor scenarios, using the same kind of sensors but at different frequencies and modalities. Heterogeneous sequences have been intentionally recorded to create a more challenging dataset, preventing domain overfitting.
Moreover, we illustrate a procedure for obtaining highly accurate ground truth in large environments combining an RTK-GPS with a Bundle Adjustment schema on the \lidar~data to obtain precise trajectories. The accuracy of our ground truth, validated with a Total Station is $\pm$3 cm along an indoor/outdoor trajectory of 1.5 km. For each dataset we provide two flavors, similar to KITTI: a training version with ground truth available and a benchmarking version where the ground truth is not publicly provided. The results of the community on the training datasets will be evaluated off-core. The public benchmark with the leading table will be available at our website.

\section{Related Work}
\label{sec:related}

\begin{figure}
\vspace{0.3cm}
     \centering
    \includegraphics[width=\linewidth]{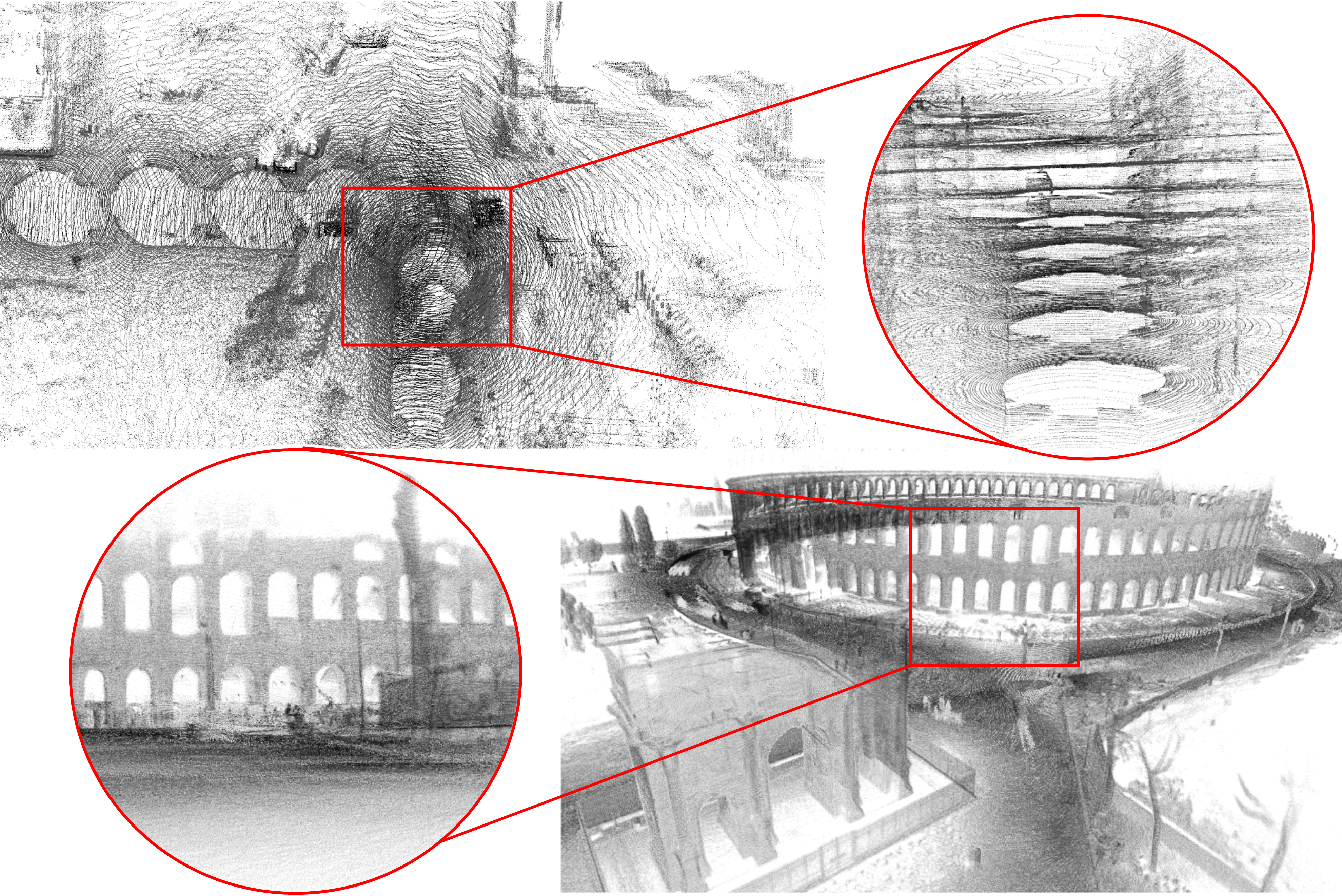}
     \caption{Comparison between \lidar~clouds attached to ground truth trajectories of KITTI (up) and ours (down). The zoom shows the elevation view.}
     \label{fig:kitti-problems-1}
\end{figure}

\begin{figure}
     \centering
     \begin{subfigure}[b]{0.99\linewidth}
         \centering
         \includegraphics[width=\linewidth]{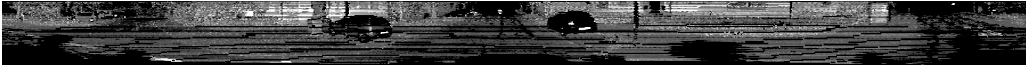}
     \end{subfigure}
     \hfill
     \begin{subfigure}[b]{0.99\linewidth}
         \centering
         \includegraphics[width=\linewidth]{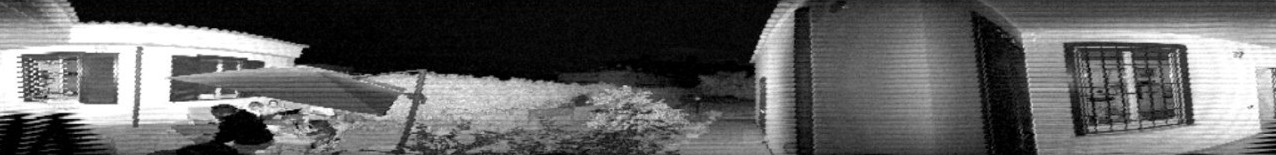}
     \end{subfigure}
     \caption{Projection of the KITTI \lidar~point cloud into an image plane (up), projection of our \lidar~ into an image plane (down). The many holes of the up image due to uneven distribution of the \lidar~beams and calibration issues make the KITTI \lidar~image unusable for computer vision tasks.}
     \label{fig:kitti-problems-2}
\end{figure}

Within the domain of SLAM and 3D reconstruction, several datasets have played pivotal roles in benchmarking and advancing autonomous robotics systems and algorithms. These datasets vary in terms of their operational contexts, sensory configurations, and the accuracy of their associated ground truth data.
One of the seminal car datasets in this field is KITTI. Its strength is providing different benchmarks (\ie~visual odometry, optical flow, stereo matching, and object detection). The KITTI datasets are publicly available and divided into training and evaluation sequences, fostering fair comparisons among the approaches.  Despite being made available for more than a decade nowadays, KITTI is the most used benchmark in robotics perception and computer vision.  
However,  KITTI presents synchronization issues between IMU readings and images, the ground truth data for visual odometry is produced only by fusing RTK GPS receiver and IMU, and the hardware used for recording data is nowadays outdated (\figref{fig:kitti-problems-1}, \figref{fig:kitti-problems-2}). Still, KITTI was pivotal in the development of many popular SLAM methods. 
Oxford RobotCar is another noteworthy car dataset \cite{maddern20171}. In contrast to KITTI, it distinguishes itself by featuring the longest sequences among the datasets. Yet, the ground truth in the Oxford RobotCar dataset relies only on partial GPS and INS data, which makes the baselines unreliable for benchmarking the accuracy of SLAM and localization methods.
Furthermore, approaches like Mulran use the same ground truth generation process, leading to the same issue. However, this dataset is renowned for embracing multimodal sensor data, including \lidar~and radar. While this adds diversity to the sequences, only the front half of the \lidar~field-of-view is included in the data collection \cite{kim2020mulran}.

In contrast to datasets collected from ground-based vehicles, certain research efforts have focused on data acquisition from micro aerial vehicles. For instance, the EuRoC dataset \cite{burri2016euroc} stands out for its use of synchronized hardware and a laser tracking system to attain accurate ground truth data. However, the dataset does not provide \lidar~aquisition but only a stereo-camera and an IMU. In addition, data is recorded only in industrial environments.

Recent advancements in handheld datasets, exemplified by Newer College \cite{ramezani2020newer} and Hilti \cite{zhang2022hilti}, have achieved exceptional levels of accuracy in ground truth generation through the use of 3D imaging laser scanners. This innovative technique involves acquiring a prior map and registering \lidar~point clouds using a localization approach. Nevertheless, a notable limitation in this case is the impracticality of applying this method to large-scale scenarios, which constrains its broader utility.

This paper introduces a diverse and heterogeneous dataset encompassing a wide range of environments. Our design accommodates various robotic platforms, including quadrupeds, quadrotors, and autonomous vehicles, making it a versatile resource for the robotics community.

We maintain hardware synchronization to ensure data accuracy and reliability and employ stereo cameras with a wide baseline to capture robust visual information. Furthermore, we provide an accurate 6-dof ground truth even in large scale scenarios.



\begin{figure}
  \centering
  \includegraphics[width=0.9\linewidth]{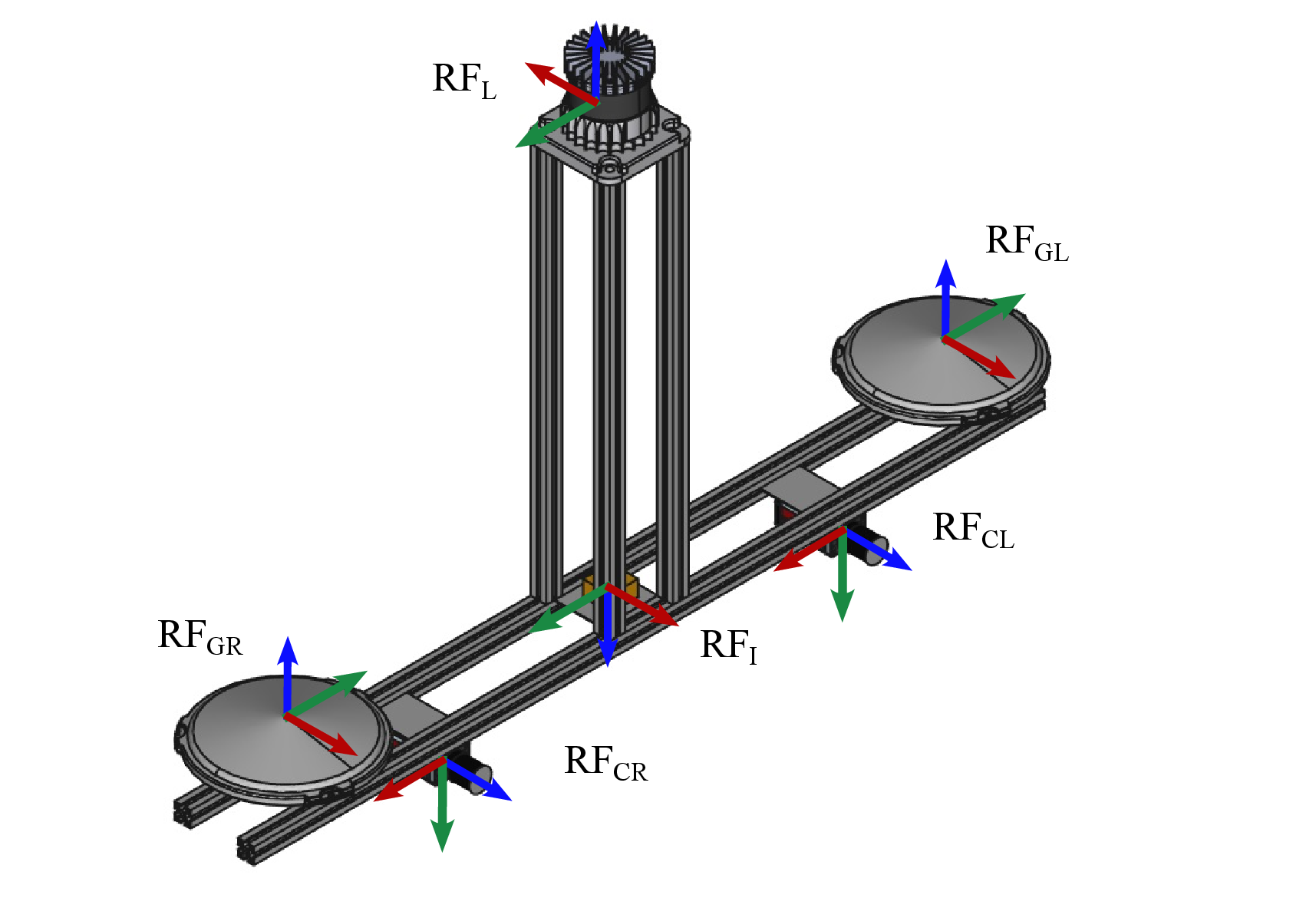}
  \caption{Sensor setup and reference frames. Our ground truth is expressed in the \lidar{} reference frame $\mathrm{RF_{L}}$. More details can be found in our website and supplementary materials.}
  \label{fig:palo}
\end{figure}

\begin{table*}
\vspace{1cm}
\centering
 \tabcolsep=0.30cm
 
\begin{tabular}{c|c|c|c|c|c|c}
\cline{2-7}
& \textbf{Accurate GT} & \textbf{Indoor} & \textbf{Outdoor} & \textbf{Various Motion} & \textbf{Large Scale} & \textbf{Benchmark} \\ \hline
\multicolumn{1}{l|}{KITTI \cite{geiger2012we}}           &                                         &                 & X                &                & X                    & X                  \\ \hline
\multicolumn{1}{l|}{EuRoC \cite{burri2016euroc}}           & X                                          & X               &                  &                  &                      &                    \\ \hline
\multicolumn{1}{l|}{Oxford RobotCar \cite{maddern20171}} &                                            &                 & X                &                 &                      &                    \\ \hline
\multicolumn{1}{l|}{ETH3D \cite{schops2019bad}}           & X                                        & X               &                  &                  &                      & X                  \\ \hline
\multicolumn{1}{l|}{MulRan \cite{kim2020mulran}}          &                                            &                 & X                &                 & X                    &                    \\ \hline
\multicolumn{1}{l|}{Newer College \cite{ramezani2020newer}}   & X                                          &                 & X                &                  &                      &                    \\ \hline
\multicolumn{1}{l|}{Hilti \cite{zhang2022hilti}}           & X                                          & X               & X                & X                 &                      &   X                 \\ \hline
\multicolumn{1}{l|}{Kitti-360 \cite{liao2022kitti360}}           & X                                          &                  & X                &                 & X                      &   X                 \\ \hline
\multicolumn{1}{l|}{\textbf{Ours}}                            & X                                       & X               & X                & X                & X                    & X                  \\ \hline
\end{tabular}
\caption{The table summarizes the most important datasets in robotics perception and computer vision related to odometry estimation and SLAM. For "Accurate GT" we mean any ground truth recorded with motion capture, Laser Total Station, or globally refined.}
\end{table*}

\section{The Datasets}
Creating comprehensive and authentic benchmarks for the tasks mentioned earlier is challenging. 
These challenges encompass collecting vast data in real-time, 
calibrating different sensors operating at various speeds, 
producing accurate ground truths with minimal oversight, 
and choosing the right sequences and frames for every benchmark. 
The following section delves into our approaches to address these issues.

\subsection{Sensors setup}
Our sensor system is illustrated in \figref{fig:palo} and consists of two  RGB cameras, a 3D LiDAR, an RTK-GPS, and an IMU.
The cameras are two global shutter Manta G-145 capturing in \rgb~and arranged in a wide stereo fashion, with a baseline of approximately 50 cm. The cameras have a horizontal FoV of $45^\circ$ and a vertical one of $40^\circ$. During the acquisition, we enable auto white-balance and auto-exposure, while maintaining a fixed focus. The maximum exposure is fixed at $20$ ms. 
Each image is $1388 \times 700$ pixels and stored in Bayer pattern to reduce the memory footprint without losing information.

Two \lidars~ were employed, tailored to the specific motion characteristics of the captured sequences. For hand-held sequences, an Ouster OS0-128 was used. This sensor offers a maximum range of 55 meters and a vertical FoV of $90^\circ$ spanned by $128$ beams. For car sequences, we used an Ouster OS1-64. This sensor provides a longer maximum range of 120 meters, a narrower vertical FoV of $45^\circ$ spanned by $64$ lasers.
 
The \gls{imu} is an SBG Ellipse-E \gls{imu}, with 0.05 $^\circ$ of roll/pitch accuracy, whose firmware supports GNSS integration.  The RTK-GPS antennas are two Septentrio PolaNt-x MF GNSS antennas mounted in differential configuration. The GPS receiver supports multi-frequency GPS, GLONASS, Galileo, BeiDou, QZSS, NavIC, Compass and L-band signal reception with an accuracy open-sky condition of 0.6 cm horizontally and 1 cm vertically. All sensors are rigidly attached to an aluminum frame. 
The relative position of the sensor is the same for both the hand-held datasets and for the driving datasets. \figref{fig:3dCar} shows the sensor placement on the car.
\tabref{tbl:sensor-setup} summarizes the devices used in our system.
\begin{table}
	\centering
 \vspace{0.1cm}
 \tabcolsep=0.15cm
  \scriptsize{
	\begin{tabular}{cccc}
    \toprule
		Sensor & Type  & Details  & Rate\\
            \hline          
		LiDARs & 
                \begin{tabular}{c} Ouster, OS0-128\\  \\ Ouster, OS1-64 \\  \end{tabular}
                &
                \begin{tabular}{c} Vertical FOV: $90^\circ$ \\ Horizontal Res: 2048 \\ Vertical FOV: $45^\circ$ \\ Horizontal Res: 1024  \end{tabular} 
                & 
                \begin{tabular}{c} 10$\mathrm{~Hz}$ \\  \\ 20$\mathrm{~Hz}$ \\  \end{tabular}
                \\
		\hline Cameras &  Manta G-125B/C&\begin{tabular}{c} Global Shutter\\Stereo configuration \\ Wide baseline\end{tabular} & \begin{tabular}{c} 20 $\mathrm{~Hz}$\\30 $\mathrm{~Hz}$ \end{tabular}  \\
		\hline  IMU & SBG Ellipse-E &\begin{tabular}{c}GNSS synchronization\\0.05$^\circ$ precision \end{tabular} & \begin{tabular}{c}100 $\mathrm{~Hz}$\end{tabular}  \\
		\hline
	\end{tabular} } 
 \caption{Summary of the devices in our sensor setup, and the accuracy of the temporal synchronization.}
 \label{tbl:sensor-setup}
  \end{table} 

\subsection{Calibration}
The accuracy of intrinsic and extrinsic sensor calibration is fundamental in achieving dependable ground truth data. Our calibration process is outlined below. 

Initially, we calibrate the stereo camera intrinsically and extrinsically. Subsequently, we determine the $\bbSE(3)$ parameters that connect the coordinate systems of the laser scanner within the right camera. Finally, we align the \lidar~/camera system with GPS/IMU reference frame.
To calibrate the camera's intrinsic and extrinsic parameters, we use an A3 checkerboard. Keeping the camera steady, we move the checkerboard and detect its corners in the calibration images. Minimizing the average reprojection error allows us to find optimal parameters for our setup \cite{bradski2000opencv}. 
Using the same target, we estimate the rigid transformation between the right camera ($\mathrm{RF_{CR}}$) and the \lidar. We achieve accurate results by minimizing plane-to-plane error \cite{giacomini2023}.
For each recorded sequence we acquire the calibration data which are public available.

Determining the relative pose between the GPS/IMU and the \lidar~relies on the sensor systems' motion, since the two devices cannot observe a common target. 
To this extent, we recover a trajectory from the \lidar/camera system using a \lidar~odometry based on point-to-plane ICP. During the process, we ensure a wide range of orientations and translations essential for addressing the minimization issue. This technique is known as \textit{hand-eye} calibration \cite{dornaika1998simultaneous}
and aims at computing the sensor offset that results in the maximum overlap between two \lidar~trajectories:
the one computed by the odometry, and the one obtained by computing the \lidar~motion from the GPS measurements (after applying the estimated offset).

\subsection{Synchronization}
\label{sec:sync}
A key challenge during acquisition involves synchronizing sensors to establish a shared temporal reference. Within our platform, two separate subsystems are at play: one comprises the \lidar~and \rgb~stereo pair, while the other includes the \gnss~receiver and \gls{imu}. In the first system, the \lidar~takes on the role of the master, generating synchronization pulses during its acquisition phase based on angle data from its encoder.
For hand-held use, the \lidar~records at $10$ Hz, and the synchronization pulse activates every $120$ degrees, resulting in a $30$ Hz signal. Moreover, in the automotive setup, the \lidar~operates at $20$ Hz, with the synchronization pulse set to trigger once per revolution.
The signal triggers the frame acquisition for both cameras, leading to sub-millisecond synchronization between the frames. Once the frames are received, their timestamp is overwritten with one of the \lidar~data packets received when the encoder was at the trigger angle. This ensures an accurate hardware synchronization between the two sensors.

In contrast, the \gnss-\gls{imu} system relies on \gls{pps} protocol for synchronization, which is directly addressed by the IMU firmware.
The streams from the two subsystems are merged together by performing an offline time synchronization to determine the difference between the internal clocks of GPS and \lidar.
We exploit the internal \lidar \gls{imu} measurements to compute the temporal shift respect to the \gls{imu}, using cross-correlation. This technique allows us to obtain a maximum of $5$\,ms error considering possible un-observable phase shift between the \gls{imu}s signals that come every $10$\,ms.
A temporal drift also affects the internal clocks of the \lidar~and GPS. In the longest sequences, we observed a maximum shift between the two clocks of about 10\,ms, which is neglectable compared to the scan/image frequency and the velocity of the sensor.

\subsection{Ground truth generation}
Generating accurate trajectories is the main objective of this work; hence, major attention was dedicated to this task. Some work produces very accurate ground truth using ICP localization within 3D Total Station reconstruction ( \cite{ramezani2020newer},  \cite{zhang2022hilti}). This is not always possible when moving in a very large environment. In such cases, a GNSS RTK system is usually used. These systems can reach an accuracy of a few centimeters, but the signal quality is not always optimal. In addition, these systems provide good global estimation, but poor locality.  

In the remainder, we assume a 3D pose $\bX \in \bbSE(3)$ be represented as a homogeneous $4 \times 4$ matrix where $\bR$ is the rotation matrix, and $\bt$ is the translation vector.
With the operator $\tTov(\bX)$, we refer to the conversion of a transform in minimal form (e.g. translation vector and unit quaternion for the orientations).

We combine the GPS priors $\bZ^\mathrm{g}_{t}\in \bbSE(3)$ with a variation of \gls{ba} formulation proposed in \cite{di2023photometric}. This allows us to combine the global accuracy of the GPS with the local precision of registration approaches. Our procedure works first by computing a \lidar~odometry that expresses the relative transform $\bZ^\mathrm{sm}_{t,t+1}$ between subsequent frames. To this extent, we use the same ICP algorithm used for temporal synchronization mentioned in \secref{sec:sync}. This odometry is accurate, reliable in the short term, and can cope with GPS outages.
Subsequently, we determine a global alignment of all the poses using the RTK-GPS readings and considering the incremental measurements of the \lidar~odometry. In short, we solve the following optimization problem:
\begin{align}
 \bX^*_{1:T} = \argmin_{\bX_{1:T}} &\sum \lVert \log( \bZ^{\mathrm{sm}-1}_{t,t+1} \bX^{-1}_{t} \bX_{t+1})  \rVert^2_{\bOmega^\mathrm{sm}_{t,t+1}} \nonumber \\
 & + \sum \lVert \log( \bZ^{\mathrm{g}-1}_t \bX_{t})  \rVert^2_{\bOmega^\mathrm{g}_{t}} \label{eq:pgo}
\end{align}
Here $\lVert \bv \rVert^{2}_\bOmega = \bv^T \bOmega \bv$ denotes the Omega L2 norm of a vector $\bv$, with  $\bOmega$ representing the information matrix encoding the accuracy of the measurement. Accordingly, 
 in \eqref{eq:pgo} $\bOmega^\mathrm{sm}_{t}$ is the information matrix resulting from scan matching, while $\bOmega^\mathrm{g}_{t}$ encodes the GPS accuracy.

Once this process is completed and we have a reasonable initial guess, we look for the set of poses that is maximally consistent with all \lidar~scans. We solve the following problem using geometric and photometric error terms. The geometric part based on point-to-plane results in a configuration of the rigid motion which is close to the optimum. The second photometric step, increases the accuracy by ensuring subpixel consistency. Let $\left<i, j\right>$ be the set of poses, $\left<k, l\right>$ geometric associations, and $u$ the image pixel generated by spherical projecting the \lidar~point cloud into an image (as illustrated in \cite{di2023photometric}). The total residual can be expressed as follows:

\begin{equation}
    E^{\mathrm{ba}} = \sum_{i,j,k,l} \rho_{\mathrm{geo}} \rVert e^{\mathrm{geo}}_{k,l} \rVert^2_{\bOmega_{\mathrm{geo}}} + \sum_{i,j,u} \rho_{\mathrm{photo}} \rVert e^{\mathrm{photo}}_{u} \rVert^2_{\bOmega_{\mathrm{photo}}}.  
\end{equation}

We employ a point-to-plane metric for the geometric error term, explicitly relying on efficient KD-tree data association based on PCA splitting criteria. The geometric term can be compactly written as: 
\begin{equation}
e_{k,l}^\mathrm{geo}(\bX_i, \bX_j) = (\bX_i \bp_{i,j} - \bX_j \bp_{k,l})\cdot (\bR_i \bn_{i,k})
\end{equation}
with $\bp_{i,k}$ and $\bp_{j,l}$ denoting corresponding points between the poses $\bX_i$ and $\bX_j$, and $\bn_{i,k}$ be the corresponding normal.
The photometric term, instead, follows the formulation illustrated in \cite{di2023photometric}, but relies only on range and intensity images. 
The overall error function to minimize, taking into account GPS information, will be therefore 
\begin{align}
\bX^{gt}_{1:T} = E^{\mathrm{ba}} + \sum \lVert \log(  \bZ^{\mathrm{g}-1}_t \bX_{t})  \rVert^2_{\bOmega^\mathrm{g}_{t}}
\label{eq:multi-view-icp}
\end{align}

We measured the accuracy of our ground truth using a Total Station and 6 highly reflective markers disposed as a hexahedron in the scene, to lock all redundantly all degrees of freedom. Since ranges are invariant of reference frame, we measure the differences between the distances measured from the points acquired from Total Station and the one detected in our estimated map. Our 6-dof ground truth results in $\pm 3$ cm accuracy on a trajectory of length of approximately 1.5 Km (indoor/outdoor).  Our ground truth generation process has been shown to scale well to large environments while relying only on the onboard sensors. The final estimated global clouds are usually in the order of billions of points.

We release the ground truth for each training sequence, always expressed in the \lidar{} reference frame $\mathrm{RF_{L}}$.  

\subsection{Data selection}

\begin{figure*}
    \centering
    \begin{subfigure}{0.8\textwidth}
        \includegraphics[width=\linewidth]{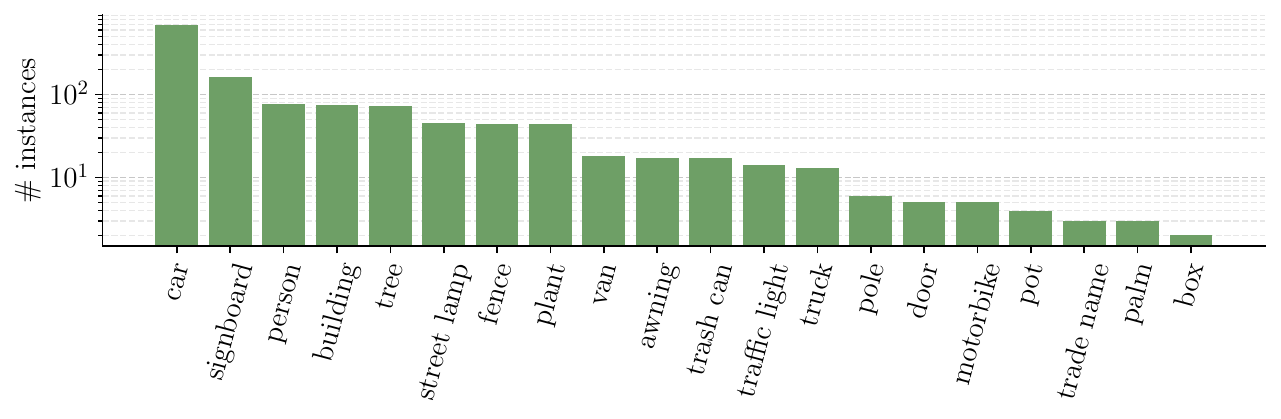}
        \label{fig:sub1}
    \end{subfigure}
    \begin{subfigure}{0.8\textwidth}
        \includegraphics[width=\linewidth]{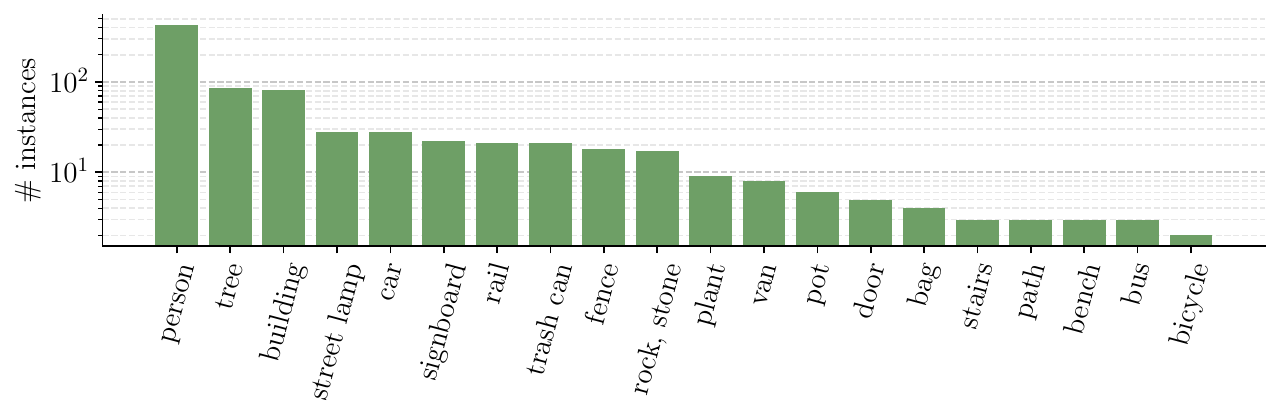}
        \label{fig:sub2}
    \end{subfigure}

    \caption{Number of top 20 most frequent semantic instance for Ciampino (above) and Colosseum (below) sequences. The instances were counted using OneFormer \cite{jain2023oneformer} over a subset of images for each sequence and excluding the most predominant classes: sky, wall, road, grass, sidewalk, ground.}
    \label{fig:semantic}
\end{figure*}
In the context of this research study, the OS0-128 \lidar~system offers extensive capabilities for collecting spatial data. Specifically, it delivers precise range measurements across the entire horizontal plane, covering large distances. Furthermore, it employs an dense array of vertical beams distributed over a $90^\circ$, enabling comprehensive scans of a spherical area surrounding the sensor.

Given the nature of the chosen environments and to ensure a rich and diverse dataset, we used various configurations provided by the \lidars~with varying resolution and frequency.
We used the OS1-64 for car sequences at 20$\mathrm{~Hz}$ to maximize the observation in wide scenarios and to reduce the skewing effect at higher speeds. Moreover, we employed the OS0-128 for hand-held sequences at 10$\mathrm{~Hz}$  to maximize the observation in narrow scenarios.


We provide 6 datasets split into different sequences. Among the datasets, 4 were acquired by walking using the hand-held device, while the other 2 collected by car. Each sequence was collected in a different environment, with different challenging scenarios such as dynamics, traffic, long sequences, and wide areas (\figref{fig:semantic}). \tabref{tbl:our-datasets} summarizes some parameters for the sequences and provides some illustration.

\begin{table}
\begin{tabular}{c|c}
\toprule
\textbf{Dataset} & \textbf{Detail}\\
\toprule
\begin{tabular}{c} 
Name:   \emph{Spagna} \\  
Motion: \emph{hand-held}\\
Type: \emph{urban/vertical}\\
Length: \emph{2.045\,km}\\
Duration: \emph{1827\,s}
\end{tabular}
& \begin{tabular}{c}\includegraphics[width=4cm]{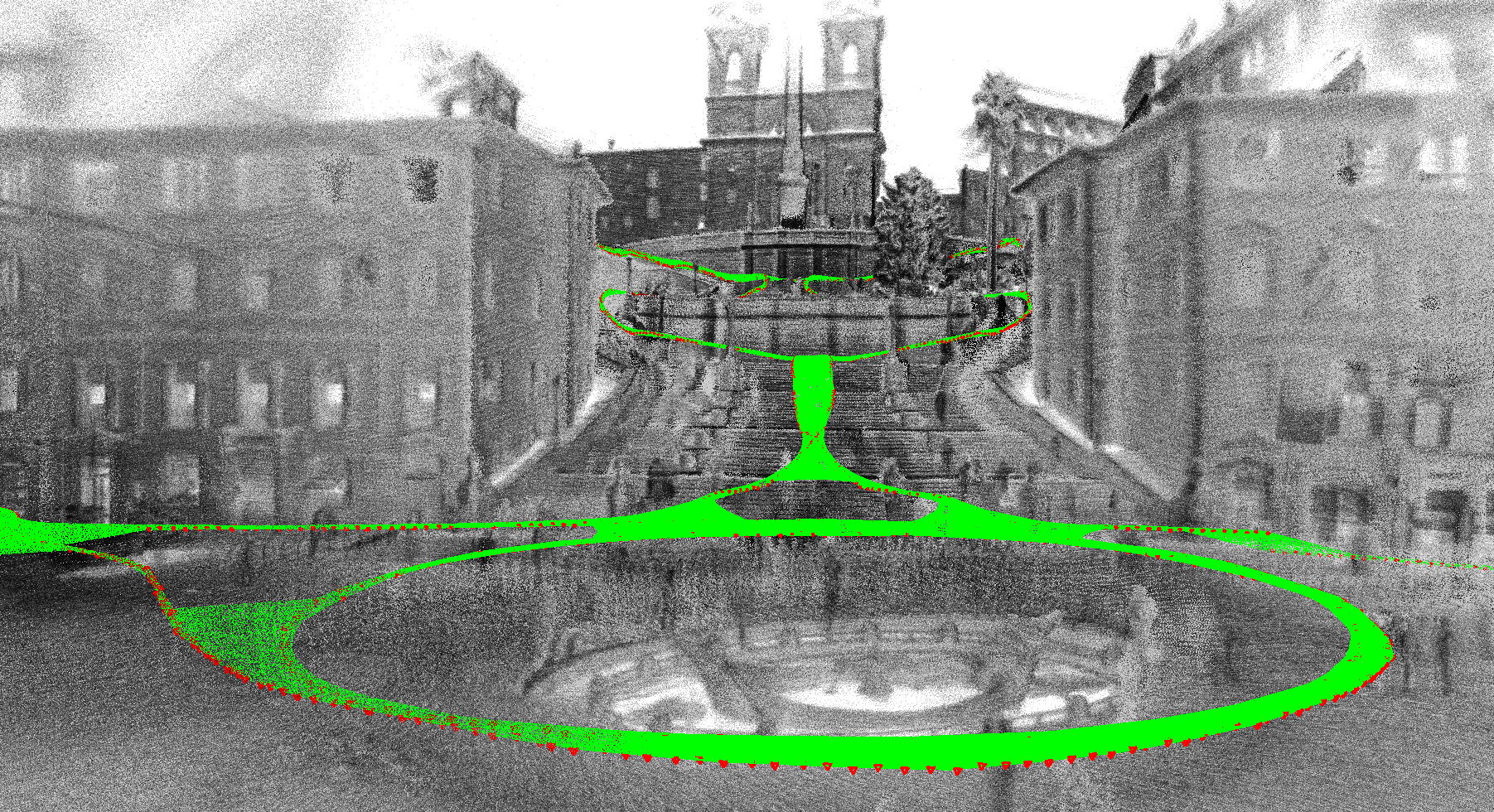} \end{tabular}\\
\hline
\begin{tabular}{c} 
Name:   \emph{Colosseum} \\  
Motion: \emph{hand-held}\\
Type: \emph{urban/dynamics}\\
Length: \emph{2.159\,km}\\
Duration: \emph{1383\,s}
\end{tabular}
& \begin{tabular}{c} \includegraphics[width=4cm]{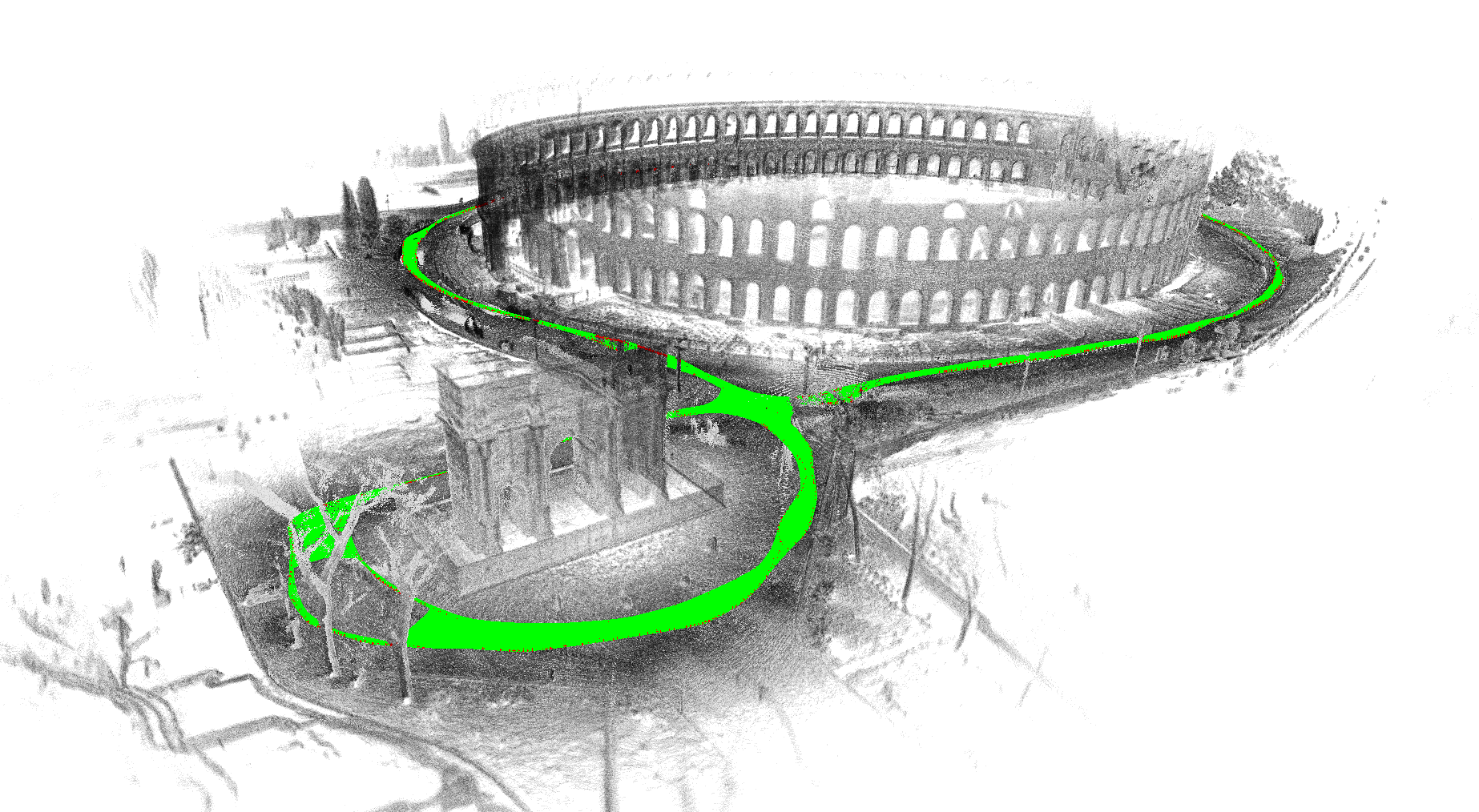} \end{tabular}\\
\hline
\begin{tabular}{c}
Name:   \emph{Pincio} \\  
Motion: \emph{hand-held}\\
Type: \emph{park/trees}\\
Length: \emph{2.541\,km}\\
Duration: \emph{2064\,s}
\end{tabular}
& \begin{tabular}{c} \includegraphics[width=4cm]{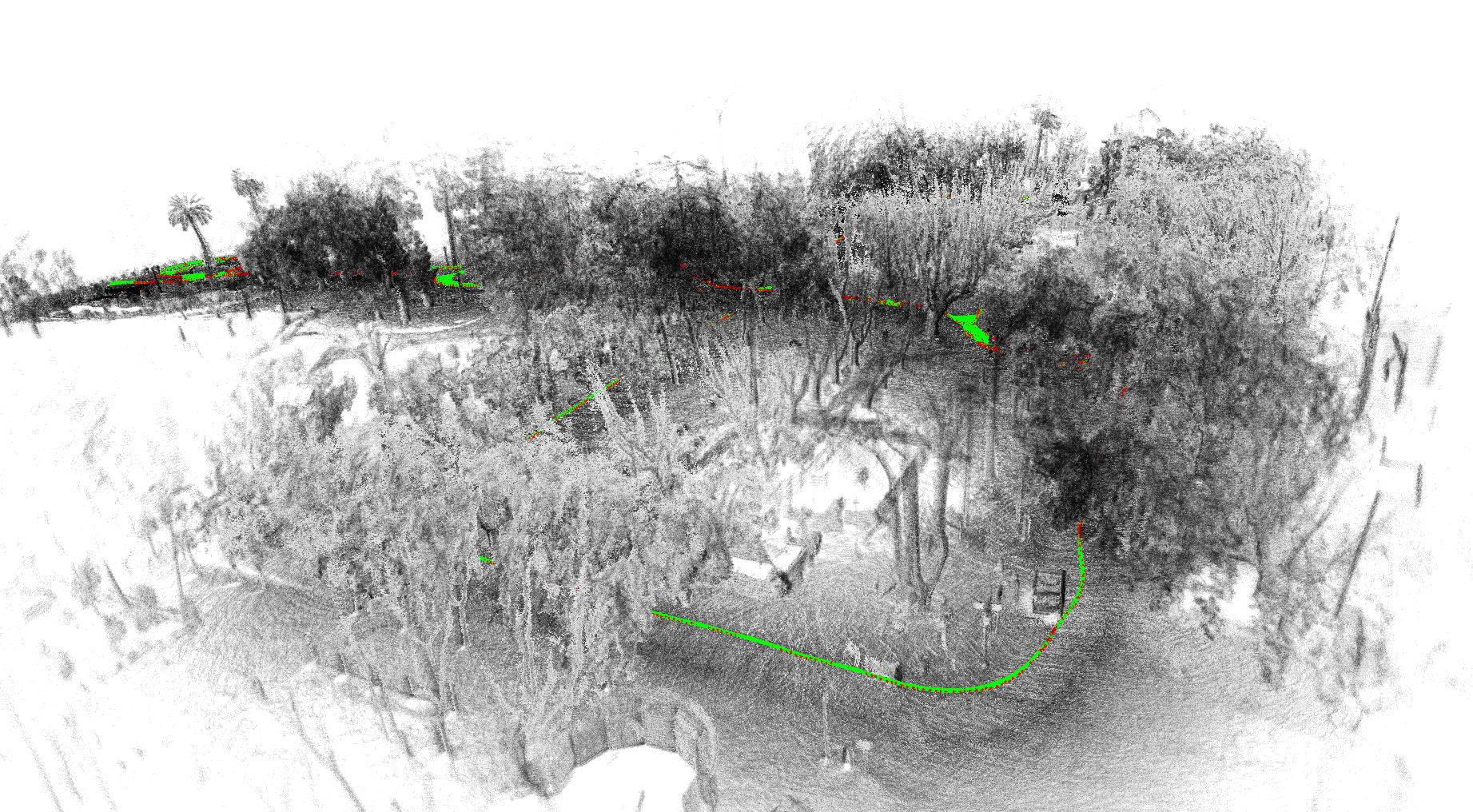} \end{tabular}\\
\hline
\begin{tabular}{c}
Name:   \emph{DIAG} \\  
Motion: \emph{hand-held}\\
Type: \emph{outdoor/indoor}\\
Length: \emph{1.480\,km}\\
Duration: \emph{1458\,s}
\end{tabular}
& \begin{tabular}{c} \includegraphics[width=4cm]{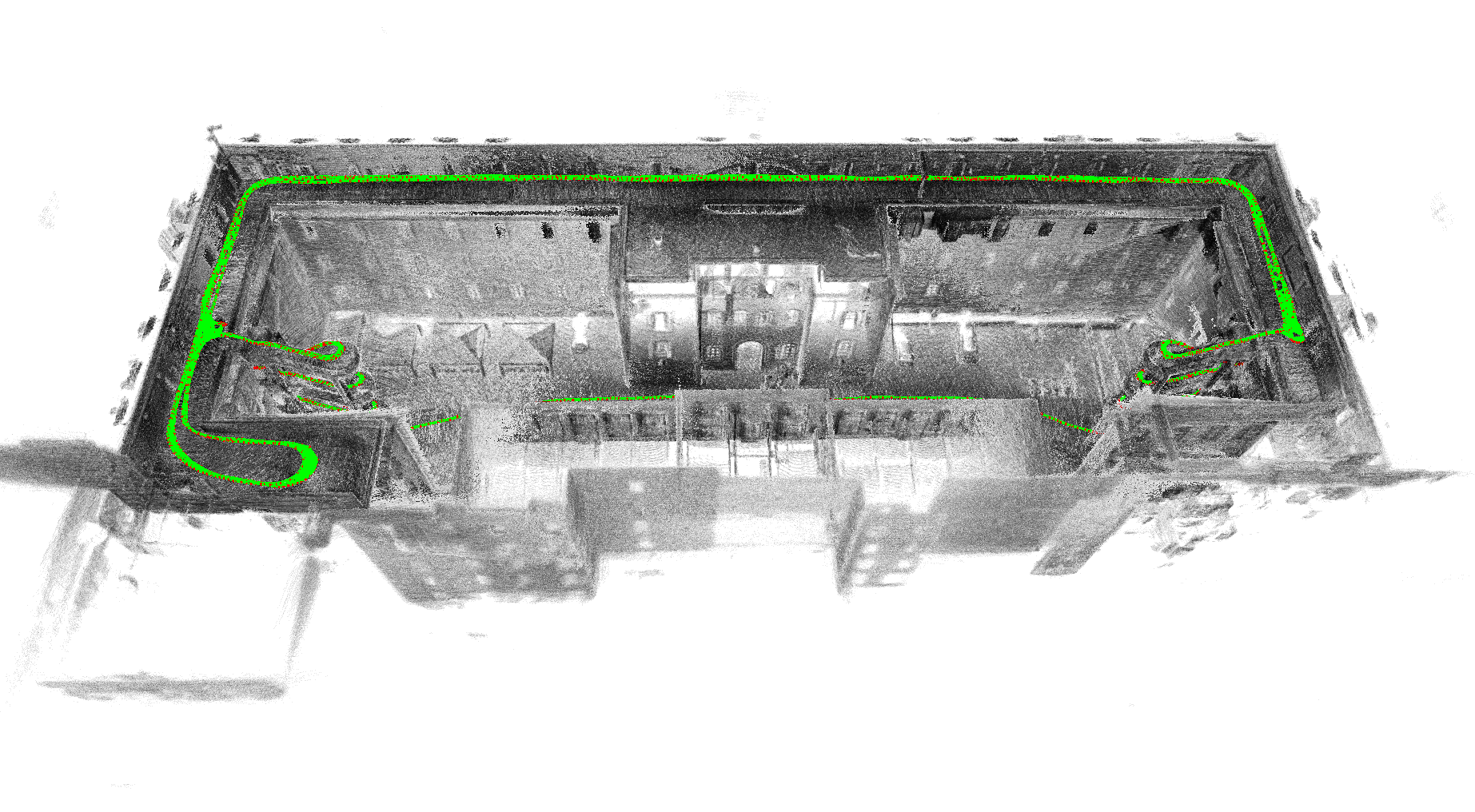} \end{tabular}\\
\hline
\begin{tabular}{c}
Name:   \emph{Campus} \\  
Motion: \emph{car}\\
Type: \emph{urban, underpasses}\\
Length: \emph{11.455\,km}\\
Duration: \emph{2290\,s}
\end{tabular}
& \begin{tabular}{c} \includegraphics[width=3cm]{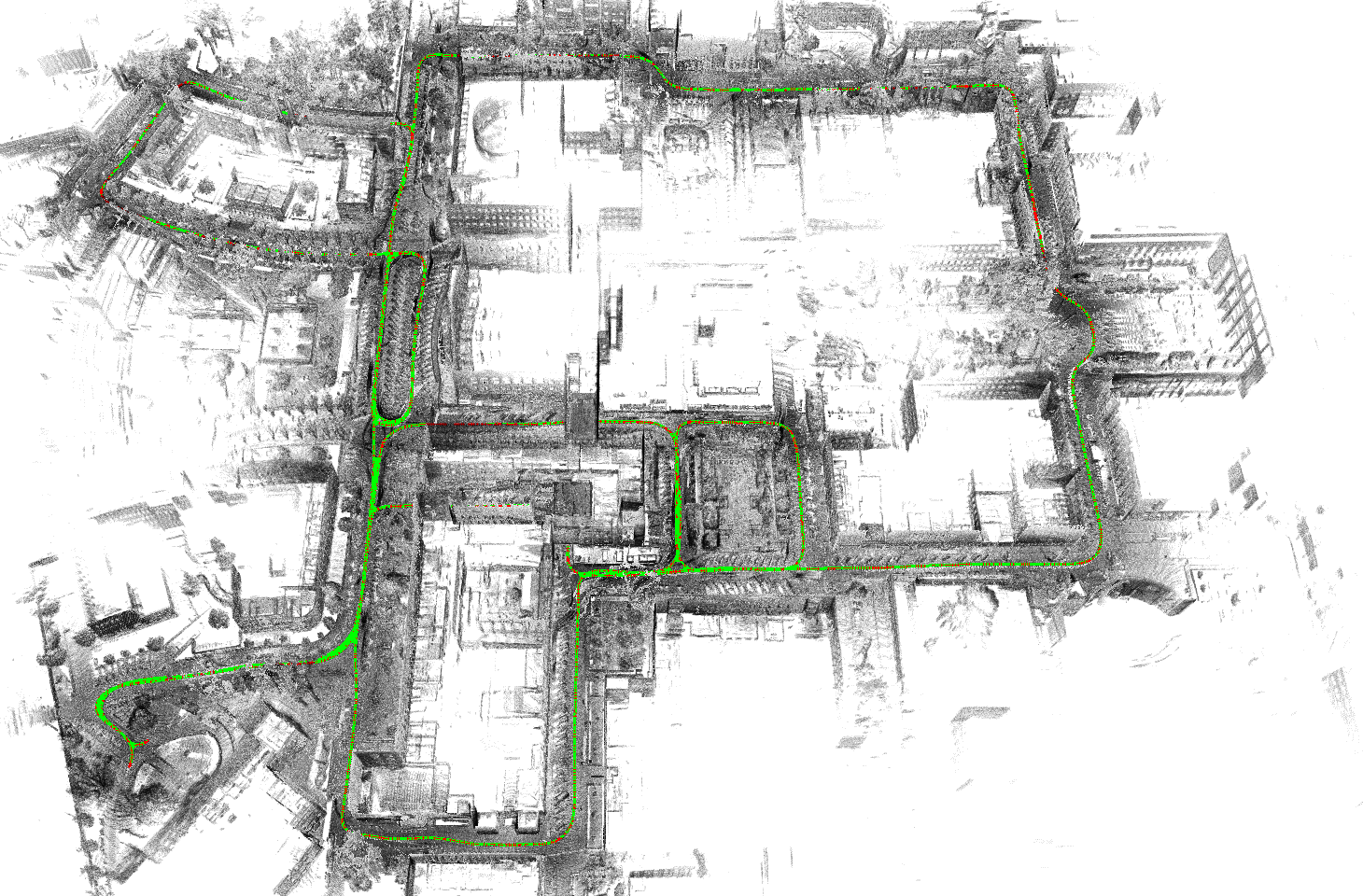} \end{tabular}\\
\hline
\begin{tabular}{c}
Name:   \emph{Ciampino} \\  
Motion: \emph{car}\\
Type: \emph{urban/traffic}\\
Length: \emph{21.064\,km}\\
Duration: \emph{3688\,s}
\end{tabular}
& \begin{tabular}{c} \includegraphics[width=4cm]{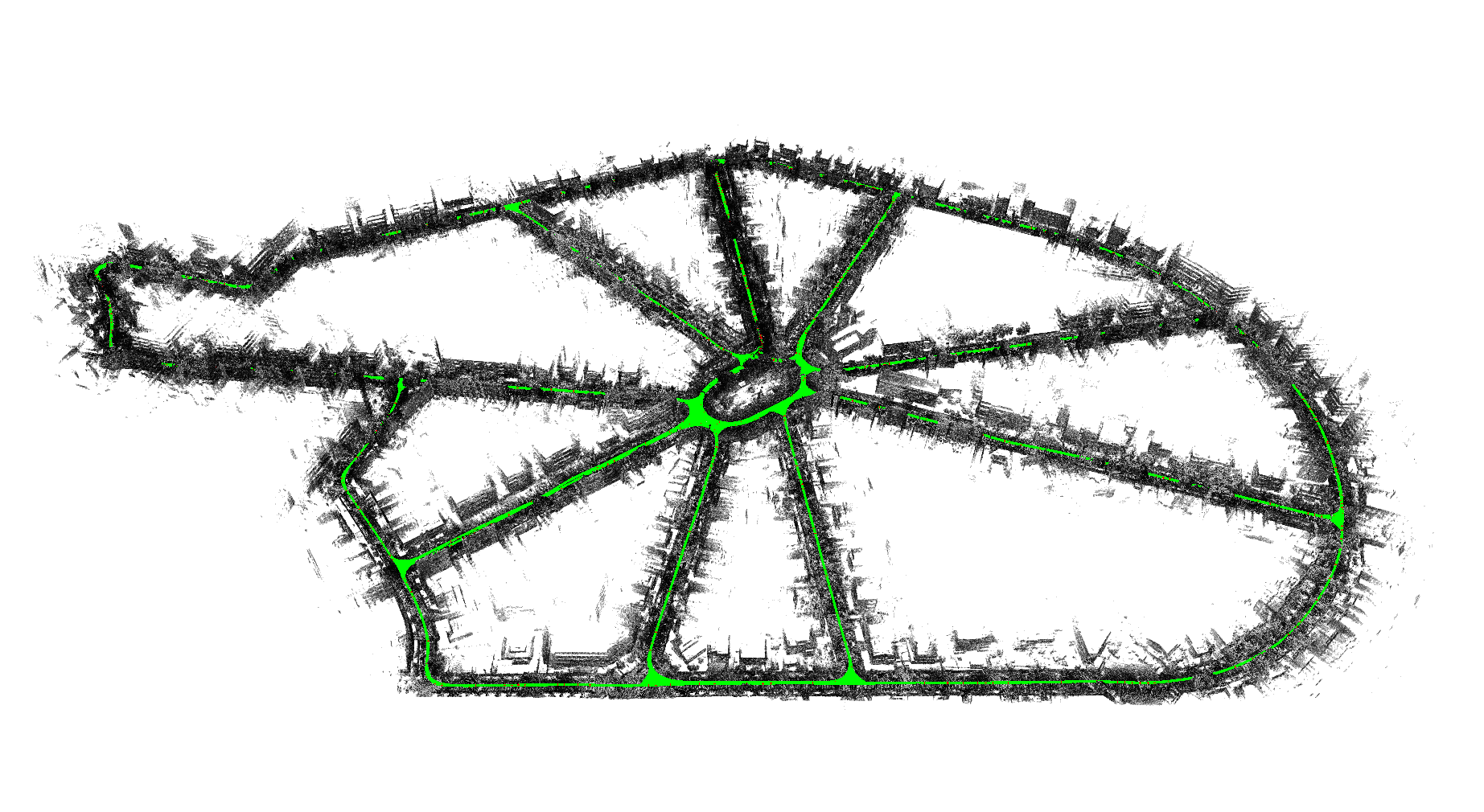} 
\end{tabular}\\
\hline
\end{tabular}

\caption{Summary of our sequences.}
\label{tbl:our-datasets}
\end{table}

In the remainder, we shortly review each sequence, describing the scenario:
\paragraph{Spagna}
this sequence has been acquired in \textit{Piazza di Spagna} and in the nearby streets. It features several large loops going up and down the stairs hence, the trajectory is non-planar. The narrow streets limit the FoV of the \lidar, but the building facades are a rich source of structure.

\paragraph{Colosseum}
consists of two rounds around the \textit{Colosseum} and the \textit{Arco di Costantino}.
The range of the \lidar~is not always sufficient to capture vertical structure. In some cases, the maximum range of the sensor is not sufficient to measure the entire surroundings, and the environment is repetitive, making some chunks difficult for state estimation.

\paragraph{Pincio}
several loops were collected in \textit{Villa Borghese}. This dataset is characterized by rich vegetation and a repetitive environment.

\paragraph{DIAG}
this sequence is a mixed indoor/outdoor dataset. We traveled inside and outside our building, walking inside the corridors, through the courtyard, up to the stairs, and on the roof, which is the only part where the reception of RTK-GPS was available.

\paragraph{Campus}
in contrast to all other sequences, this has been acquired at the main Campus of Sapienza University using the equipped car, and features several loops, spanning approximately all the streets that can be traversed by car. There are several narrow passage and some tunnels that pass under buildings. The dynamic is composed mainly of people walking composing a low percentage of the recorded data.

\begin{figure}
 \vspace{0.2cm}
  \centering
  \includegraphics[width=0.45\textwidth]{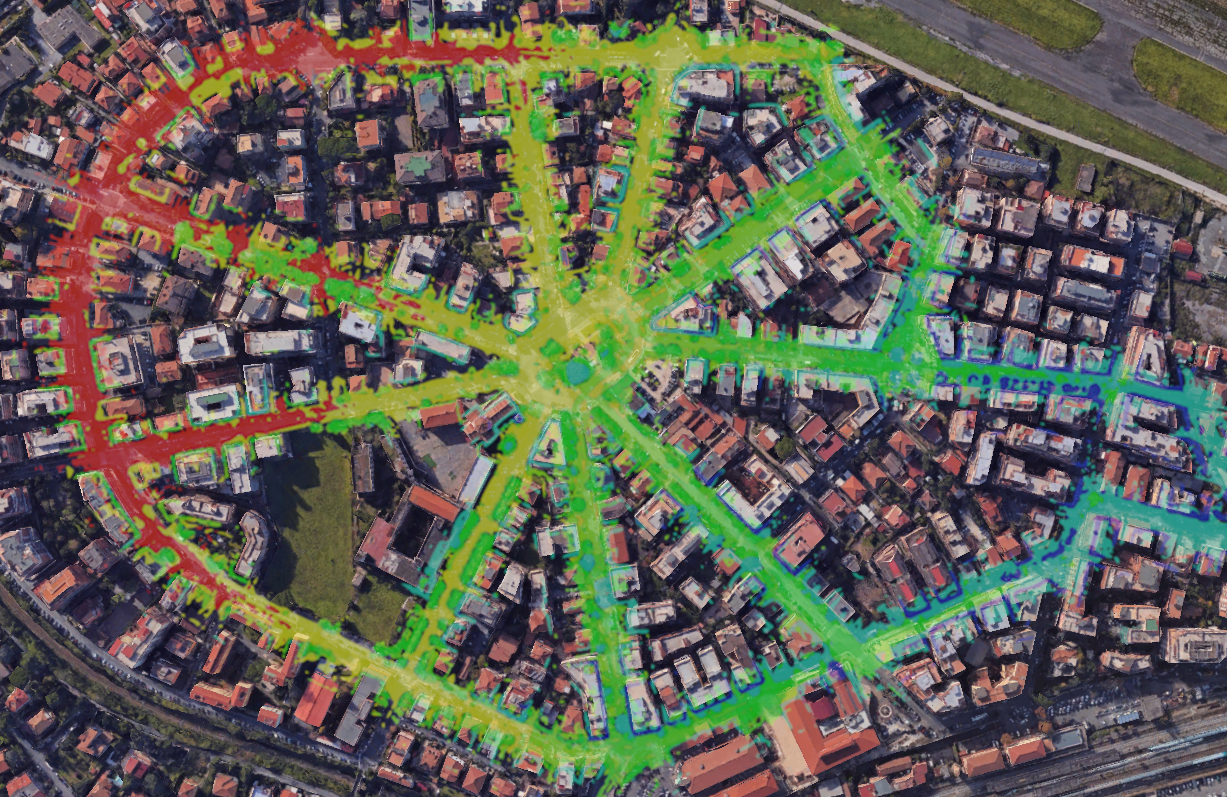}
  \caption{The image shows the overlay of the 3D model obtained from our ground truth system and a view from Google Maps.}
  \label{fig:CiampinoOverlay}

\end{figure}

\paragraph{Ciampino}
these sequences have been recorded in the city of Ciampino (\figref{fig:CiampinoOverlay}). It is the longest sequence so far: the length of the total trajectory is about 21 km, while being subject to moderate dynamics.

\section{Benchmark}
For a detailed assessment of SLAM and odometry estimation, we concentrate on the Absolute Trajectory Error (ATE) and Relative Pose Error (RPE). As in \cite{geiger2012we}, we assess rotation and translation errors independently rather than merging them into one metric.

ATE emphasizes SLAM performance over odometry. To compute its RMSE, we first align the estimated trajectory with the ground truth using a $\bbSE(3)$ transformation, matching poses with synchronized timestamps and employing the Horn method \cite{Horn:88}. Subsequently, we calculate the RMSE of the ATE [m] among all the matched poses.

On the other hand, RPE emphasizes the odometry comparing local motion estimate chunks within the ground truth. It involves computing the RPE [\%] (measured in percentage), over a set of subsequences of different lengths, as proposed by \cite{geiger2012we}. Afterwards, the translational RPE [\%] and the rotational RPE [deg/m] of a sequence are computed as the average of all chunks RPE.
Differently than any other benchmark, we have chosen to make chunk lengths adaptive to the total sequence length. In fact, local and global accuracy of our ground truth trajectories allows us to choose subsequences of arbitrary lengths, without biasing evaluation results.


Given a trajectory estimate for each sequence, a cumulative error curve is computed, like those in \figref{fig:plot-cumulative}. For a given error value on the $x$-axis, the $y$-axis shows in how many sequences a method achieves a lower error.
Therefore, the method ranking is determined as the area under curve, up to the selected maximum error and, for this metric, the larger the better. This metric rewards the robustness of evaluated methods, since a successful result on a sequence usually adds much more area under the curve than slightly improving the accuracy on many sequences.

Additional information, supplementary materials, and the leading table can be accessed on our website.

\subsection{Evaluation}
\begin{figure}
     \centering
     \begin{subfigure}[b]{0.9\linewidth}
         \centering
         \includegraphics[width=\linewidth]{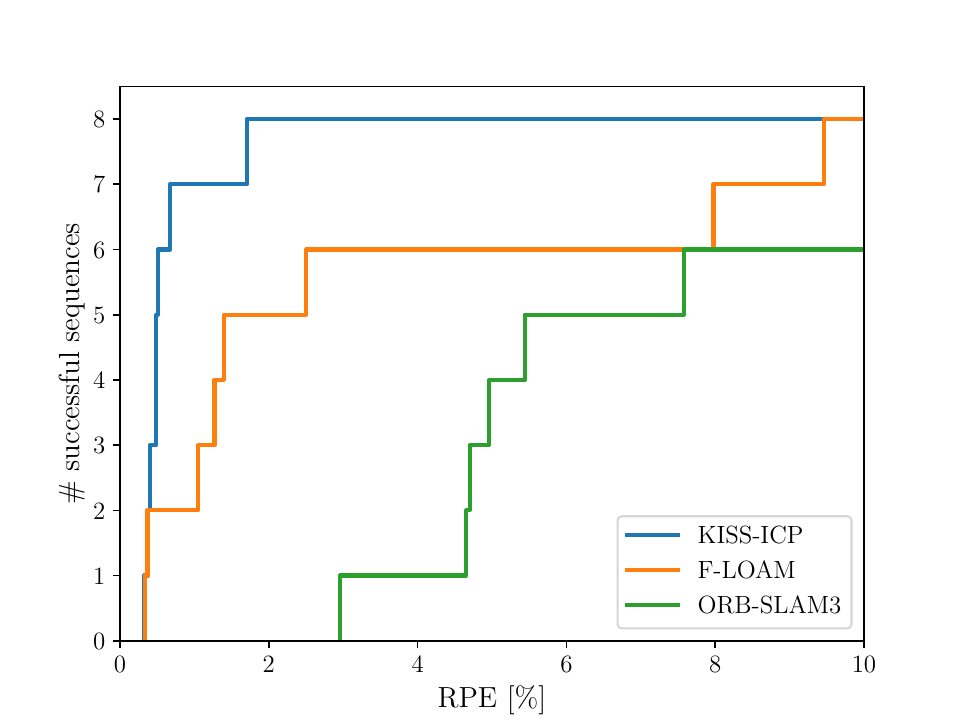}
     \end{subfigure}
     \hfill
     \begin{subfigure}[b]{0.9\linewidth}
         \centering
         \includegraphics[width=\linewidth]{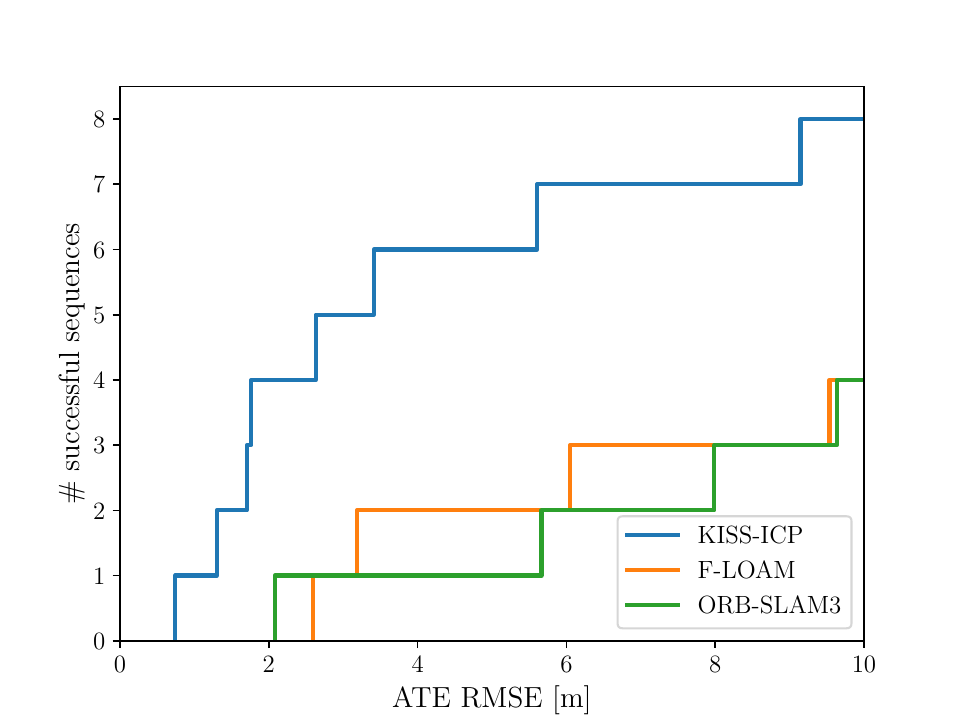}
     \end{subfigure}
     \caption{Our benchmark. Cumulative RPE [\%] and ATE RMSE [m] across all training sequences in our dataset for KISS-ICP, F-LOAM and ORB-SLAM3.}
     \label{fig:plot-cumulative}
\end{figure}

To assess our recorded data's integrity, we evaluated different \lidar~odometry and visual SLAM systems on all our training sequences, specifically focusing on three notable solutions: KISS-ICP \cite{vizzo2023kiss}, F-LOAM \cite{floam} and ORB-SLAM3 \cite{campos2021orb}.  
The evaluation outcomes are reported in \figref{fig:plot-cumulative}, showing cumulative RPE [\%] and ATE RMSE [m], with threshold limits set to 10 \% and 10 m respectively. As expected, \lidar~odometry methods are more robust and accurate than visual SLAM, in fact both KISS-ICP and F-LOAM perform successfully across all our 8 training sequences. The outcomes slightly change in the ATE RMSE [m] graph, where the area under the curve of every method is reduced, emphasizing the challenges in estimating the egomotion with global accuracy.

\section{Conclusion}
\label{sec:conclusion}
In this paper, we present a new vision and perception dataset, specifically targeted at SLAM and odometry estimation methods.  Our sequences cover different environments and are acquired in a hand-held fashion and by using a car. 
Our design is to accommodate various types of robotic platforms,
including quadrupeds, quadrotors, and autonomous vehicles,
making it a versatile resource for the robotics and vision community. Compared to existing datasets, we offer a variety of environments within our sequences. Moreover, this work presents a novel ground truth estimation, fusing an RTK-GPS with a \lidar~Bundle Adjustment schema. All the sequences are split into training and test sets. In addition we provide a public benchmark evaluation system, accessible from our website, that produces a leading table from the results submitted by the community. 

As a further service to the community, we plan to extend our benchmark with other sequences, annotations and challenges in the area of computer vision and robotic perception (\ie~semantics, monocular and stereo dense depth estimation, object tracking, etc.). 

\section*{Acknowledgement}
We thank Juan D. Tardos for supporting us with ORB-SLAM. We are grateful to Roberto Mauroni and Francesco Spognardi for their invaluable support. Special thanks to Vincenzo Suriani for providing supplementary hardware. Last but not least, we appreciate Luca Callocchia's contribution in developing the website.

\bibliographystyle{abbrv}

\bibliography{2023_paper_giacobrizi_endless_icra.bib}

\end{document}